\documentclass[
]{ceurart}

\usepackage{listings}
\usepackage{wrapfig}
\usepackage{hyperref}
\hypersetup{colorlinks=true, linkcolor=blue}
\newif\ifcomment
\commenttrue 

\begin{document}

\copyrightyear{2025}
\copyrightclause{Copyright for this paper by its authors.
  Use permitted under Creative Commons License Attribution 4.0
  International (CC BY 4.0).}

\conference{ECOM'25: SIGIR Workshop on eCommerce, Jul 17, 2025, Padua, Italy}

\title{S2SRec2: Set‑to‑Set Recommendation for Basket Completion with Recipe}


\author[1]{Yanan Cao}[%
email=yanan.cao@walmart.com,
]
\address[1]{Walmart Global Tech, Sunnyvale, CA, 94086, United States}

\author[1]{Omid Memarrast}[%
email=omid.memarrast@walmart.com,
]

\author[1]{Shiqin Cai}[%
email=shiqin.cai@walmart.com,
]

\author[1]{Sinduja Subramaniam}[%
email=sinduja.subramaniam@walmart.com,
]

\author[1]{Evren Korpeoglu}[%
email=evren.korpeoglu@walmart.com,
]

\author[1]{Kannan Achan}[%
email=kannan.achan@walmart.com,
]


\begin{abstract}
In grocery e-commerce, customers often build baskets of ingredients guided by dietary preference, but lack the recipe expertise to create complete meals. Consequently, utilizing recipe knowledge to recommend complementary ingredients based on given ingredients is key to filling these gaps for a successful culinary experience. The traditional method for completing a given ingredient set, also known as recipe completion, generally focuses on predicting a single missing ingredient using a leave-one-out strategy from recipe data; however, this approach falls short in two important aspects when applied to real-world scenarios. First, these methods do not fully capture the complexity of real-life culinary experiences, where customers routinely need to add multiple ingredients to complete a recipe. Second, they only consider the interaction between the existing ingredients and the missing ingredients but neglect the relationship among multiple missing ingredients. To overcome these limitations, we reformulate basket completion as a set-to-set (S2S) recommendation problem, where an incomplete basket is input into a system that predicts a set of complementary ingredients to form a coherent culinary experience. We introduce S2SRec2 a set-to-set ingredient recommendation framework utilizing a Set Transformer based model trained in a multitask learning paradigm. S2SRec2 simultaneously learns to: (i) query missing ingredients based on the set representation of existing ingredients, and (ii) determine the completeness of the basket based on the union set of existing and predicted ingredients. These two tasks are jointly optimized, which enforces both accurate retrieval of complementary ingredients and coherent basket completeness prediction for multi-ingredient recommendation. Experiments on large-scale culinary datasets, together with extensive qualitative analyses, demonstrate that S2SRec2 significantly outperforms traditional single-target recommendation methods, offering a promising solution to enhance grocery shopping experiences and foster culinary creativity.
\end{abstract}





\maketitle

\section{Introduction}
\begin{figure*}[htbp]
\centering
\includegraphics[width=0.9\linewidth]{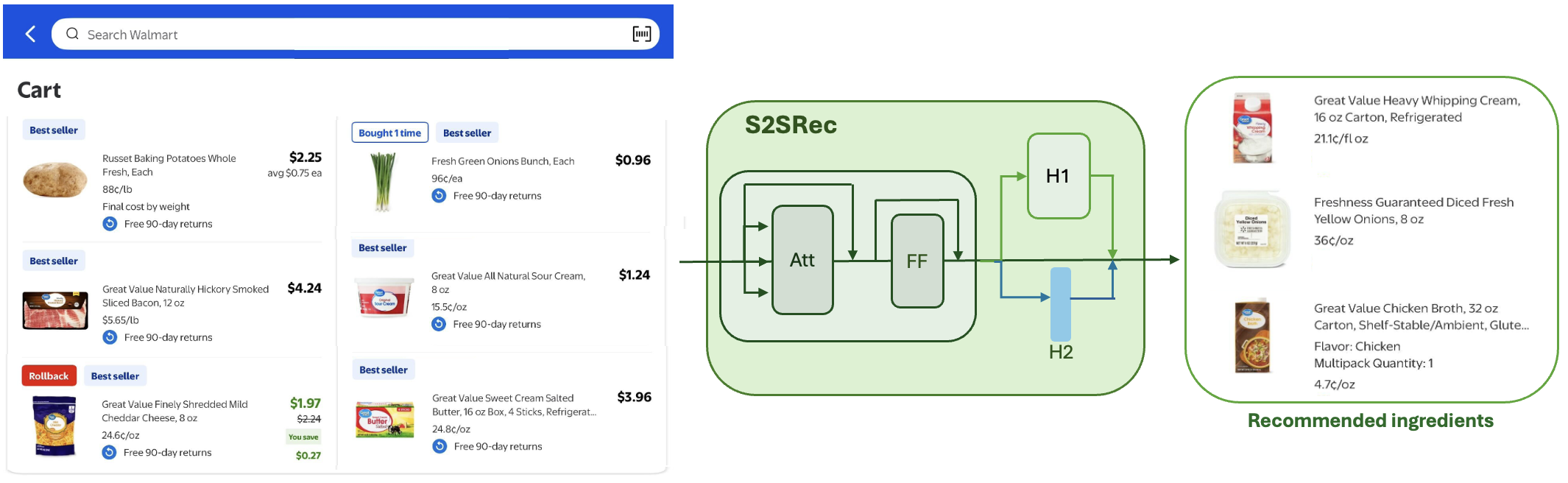}
\caption{Demonstration of recipe completion for a grocery basket. Given a basket with ingredients, S2SRec2 utilizes the knowledge of recipe ingredient to generate recommended ingredients that can form a recipe. In this example, given potato, bacon, cheese, green onion, sour cream and butter, S2SRec2 recommends heavy cream, yellow onion and chicken broth for cooking a potato soup.}
\label{fig:basket_demo}
\end{figure*}

\vspace{-2mm}
The convergence of machine learning and culinary arts has sparked growing interest in understanding recipes through textual data, such as ingredient lists, cooking instructions and visual content like recipe images \cite{salvador2017learning}. Although there are numerous recipes available, recipe completion – identifying the right set of potential ingredients that complement those already present - remains a challenging yet crucial problem. In grocery e-commerce, recipe completion aligns naturally with the task of basket completion. Ingredient recommendations for completing recipes with partial ingredients in an e-commerce basket can not only streamline a user’s meal planning process and reduce food waste but also improve the grocery shopping experience, leading to higher customer satisfaction and conversion metrics for business. Figure \ref{fig:basket_demo} demonstrates an example use case of recipe completion for a grocery basket: if a customer has a set of ingredients whose variety is not sufficient for cooking, a recipe completion system can suggest potential grocery items that can form a decent dish with existing ingredients. By predicting missing ingredient and suggesting additions, these systems facilitate efficient meal preparation and ensure informed and satisfactory choices during grocery shopping. 
There are a handful of studies that have explored methods for completing partial recipes in different ways, but each has its own limitations. Traditional methods such as item-based collaborative filtering, matrix factorization \cite{de2016data}, and content-based filtering \cite{nilesh2019recommendation} are commonly used in the early works of recipe completion and recipe-driven food recommendation. However, such methods neglect the rich patterns in the recipe data, such as the compatibility among ingredients and the relationship between ingredients and recipes. To leverage the rich information in the recipe data, recent studies in this field have started exploring Knowledge Graph (KG) \cite{guo2022research, tian2022recipe2vec} and Set-Transformers \cite{li2020reciptor, gim2021recipebowl, gim2022recipemind} for recipe representation learning to facilitate the recipe completion task. However, for recipe completion, the Leave-One-Out (LOO) \cite{li2020reciptor, gim2021recipebowl, gim2022recipemind} method is commonly used to predict the single target ingredient excluded from the original set of ingredients. Such approaches overlook the fact that most real-life scenarios require multiple ingredients to complete a recipe.

\vspace{-2mm}

To address the limitations in previous works, we propose Set-to-Set Recommendation for Recipe completion (S2SRec2), which takes a set of existing ingredients as input and predicts a set of potential ingredients that complement the input set of ingredients to form a viable recipe prediction with a higher level of confidence. S2SRec2 is a Set Transformer based model trained in a multi-task learning fashion. In S2SRec2, we leverage the inherent permutation invariance of the Set Transformer to represent a recipe as an unordered set of ingredients, thereby capturing complex inter-dependencies without being affected by the order of elements.
Rather than predicting a single missing ingredient, S2SRec2 is formulated as a set-to-set recommendation problem in which the system is tasked with proposing the missing ingredients and predicting the completeness of the basket when combined with the predicted ingredients. To achieve this, our model is trained under a multitask learning framework that simultaneously optimizes two objectives: (i) a missing-ingredient query task, where a learnable query attends to the encoded ingredient representations to predict which ingredient should be added to the current basket, and (ii) a recipe completeness prediction task, which indicates whether the basket with additional ingredients is complete. These tasks are jointly optimized, ensuring both precise ingredient retrieval and reliable detection of basket completeness in multi-ingredient recommendations. Our main contributions are highlighted as follows:
\vspace{-1mm}
\begin{itemize}
\item \textbf{Set-to-set formulation for recipe completion:} To our knowledge, this is the first work to formulate recipe basket completion as a set-to-set recommendation problem. Instead of predicting a single missing item, the system recommends a set of ingredients to complement an existing ingredient set, addressing the more realistic scenario of multi-ingredient additions.
\item \textbf{Set-Transformer model with multitask learning:} We propose \textbf{S2SRec2}, a Set Transformer–based model that simultaneously learns a missing-ingredient query function and a stop signal indicating basket completeness. This design exploits the permutation invariance of sets to model ingredient combinations and uses a multi-task objective to add ingredients one by one and decide when no further items are needed.
\item \textbf{Empirical gains on large-scale datasets:} Through extensive experiments on large-scale recipe datasets, we demonstrate that \textbf{S2SRec2} outperforms existing methods (including single-ingredient recommendation approaches) in both recommending relevant ingredients and correctly predicting when the recipe is complete. \textbf{S2SRec2} delivers more coherent multi-ingredient recommendations, highlighting its effectiveness for real-world recipe completion scenarios.
\end{itemize}
\vspace{-2mm}

\vspace{-2mm}
\section{Related Work}
\vspace{-1mm}
Previous work on recommendation in the food domain mainly focuses on Recipe Completion \cite{cueto2020completingpartialrecipesusing, guo2022research, gim2021recipebowl}, Food Recommendation \cite{gim2022recipemind, wang2021market2dish}, and Recipe Recommendation\cite{tian2022reciperecheterogeneousgraphlearning, li2023health, Neelam_2024}. Since they all rely on the recipe information, including ingredients and instructions, to derive hidden patterns among ingredients and recipes, Recipe Representation Learning becomes a fundamental research field in the food domain. In this section, we review related works in Recipe Representation Learning and Recipe Completion.

\vspace{-2mm}
\subsection{Recipe Representation Learning}
\vspace{-1mm}

To effectively encode recipes into meaningful embeddings, various techniques such as Multi-modal embeddings \cite{tian2022recipe2vec}, Graph Neural Networks \cite{tian2022recipe2vec}, Knowledge Graphs \cite{guo2022research}, and Set Transformers \cite{li2020reciptor, gim2021recipebowl, gim2022recipemind} have been explored to represent recipes in a continuous vector space. A recipe can contain textual information including ingredients, cooking instructions, and tags, as well as visual information such as food images. Tian et al. \cite{tian2022recipe2vec} proposed a multi-modal recipe representation learning using a Graph Neural Network model to incorporate textual, visual, and relational information into recipe embeddings. Since the ingredients in a recipe can be represented as an unordered set of ingredients, the permutation invariant property in Set Transformers \cite{lee2019set} inspires the research in variant set transformer architectures. Li \& Zaki’s work \cite{li2020reciptor} proposes a Set Transformer-based joint model to learn recipe representations through a recipe Knowledge Graph and optimize the learned embeddings using a triplet loss to ensure similar recipes are closer in the latent semantic space. 

\vspace{-2mm}
\subsection{Recipe Completion}
\vspace{-1mm}

Although recipe representation learning has received considerable attention, studies on recipe completion remain relatively sparse. A few pioneering works have explored tackling this problem using collaborative filtering, knowledge graphs, and food pairing. In collaborative filtering-based methods, Cueto et al. \cite{cueto2020completingpartialrecipesusing} use an item-based recommender system for ingredient recommendation based on the similarity between item vectors containing their ratings. Nevertheless, such approaches neglect the rich patterns in food data, such as the interaction between ingredients and the relevance between ingredients and recipes. Guo et al. \cite{guo2022research} constructed a collaborated knowledge graph which combines a food knowledge graph of ingredients and recipes with user interaction information for food recommendation. Gim et al. \cite{gim2021recipebowl} train a supervised learning model to predict the eliminated ingredient given a leave-one-out ingredient set for each recipe, while  Kim et al. \cite{gim2022recipemind} propose a solution based on the theory of food pairing by predicting the affinity scores that quantify the suitability of adding a single ingredient to the existing set of ingredients. These works recommend only one additional ingredient, while ignoring the fact that most recipe completing scenarios need multiple additions. They also overlook the relationships among the potential ingredients, where an additional ingredient would impact the next ingredient prediction.

\vspace{-2mm}
\section{Method}
\vspace{-1mm}
In this section, we introduce S2SRec2, a set-to-set recommendation method for recipe completion that can be applied to e-commerce complementary item recommendation for grocery baskets.
In our setting, only ingredients appear in both the basket data and the recipe data, so we leave out extra recipe details such as tags, descriptions, and cooking steps used in prior work \cite{li2020reciptor, gim2021recipebowl, gim2022recipemind}.

\begin{figure*}[htbp]
\centering
\includegraphics[width=0.9\linewidth]{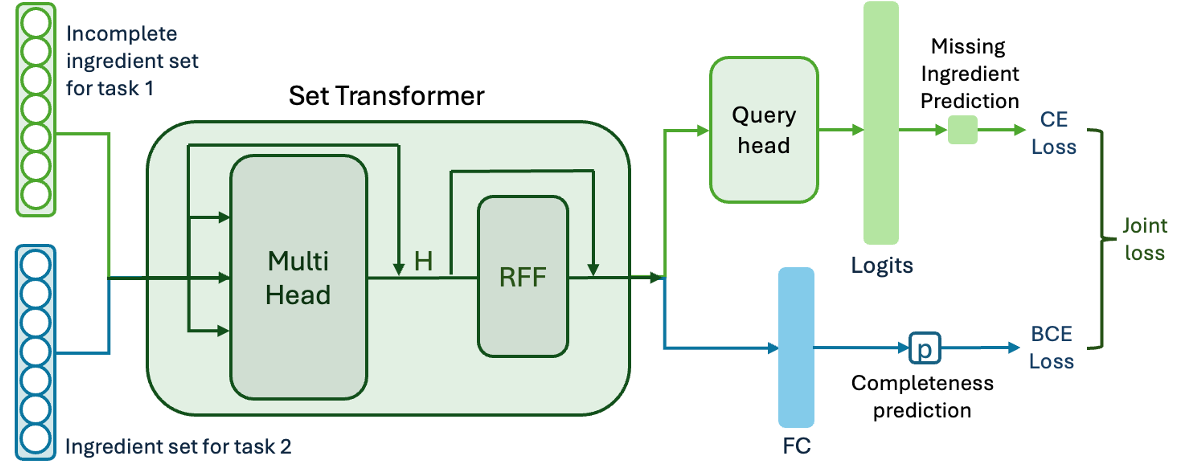}
\caption{
S2SRec2 training pipeline. The model encodes the ingredient set, predicts the missing ingredient and the basket completeness, and updates weights with a joint loss.
}
\label{fig:training}
\end{figure*}

Based on the study of related work, we adopt the Set Transformer architecture in Li \& Zaki’s work to learn the representations of a set of ingredients, considering the unordered nature of recipe ingredients. The Set Transformer consists of an encoder and a decoder, where the encoder processes each element independently by performing self-attention among the elements to generate each element's representation enriched by other elements in the set, and the decoder summarizes the set via pooling for downstream tasks. The key MAB, ISAB, and PMA blocks we use are outlined in Appendix \ref{appendix}. After the Set Transformer learning the permutation-invariant relationship between the ingredients, the model is jointly optimized on two tasks—missing ingredient prediction and completeness prediction. Figure \ref{fig:training} demonstrates the training procedure of S2SRec2.

\vspace{-5mm}

\subsection{Predicting Missing Ingredients Using a Learnable Query}
The first task of S2SRec2 is to identify the missing ingredient. We reformulate the problem as a missing-ingredient query task and introduce a learnable query that directly interrogates the encoded set representation of the existing ingredients.
Specifically, given an input basket represented as an unordered set of ingredients, a Set Transformer encoder is applied to capture the rich inter-dependencies among the ingredients without imposing any sequential order. A learnable query vector is then used to attend to the set of ingredient embeddings, effectively asking, ``which ingredient is missing to complete this basket?'' The result is a score assigned to every candidate ingredient.

For each recipe \(r_i \in R\), a subset of ingredients is randomly selected to form an incomplete basket, denoted as \(S\). Let the encoded ingredient set be denoted by \(S = \{s_1, s_2, \ldots, s_n\}\) and let \(q\) be the learnable query vector. For each ingredient \(c_i \in \mathcal{C}\), where \(\mathcal{C}\) is the set of all ingredients available for recommendation, an attention mechanism is used to compute scores between \(q\) and \(c_i\). These scores are then passed through a softmax layer to generate a probability distribution over all candidate ingredients:
\begin{equation}
P(c_i \mid S, q) = \text{softmax}(\text{Score}(q, c_i)), \quad \forall \, c_{i} \in \mathcal{C}.
\end{equation}
The ingredient with the highest probability is selected as the missing ingredient complement.

This learnable query mechanism is optimized end-to-end with a cross-entropy loss between the predicted probability distribution and the ground truth missing ingredient. 
\vspace{-2mm}
\begin{equation}
L_{\text{CE}} = -\frac{1}{N} \sum_{i=1}^{N} c_i \log\left(P(c_i \mid S, q)\right),
\end{equation}
\vspace{-3mm}

This design not only supports the recommendation of multiple ingredients sequentially as needed in real-world scenarios but also inherently models the internal interactions among the predicted missing ingredients - ensuring the recommended set forms a coherent culinary experience.

While the query head can propose ingredients one by one, the model must also decide when further additions are no longer beneficial—a capability provided by the completeness head introduced next. 
\vspace{-1mm}
\subsection{Basket Completeness Prediction using Classification}
\vspace{-1mm}

To further enhance the model's ability to recognize a successful culinary completion, we define the second task to predict the completeness of a basket formed by combining the existing ingredients with a predicted missing ingredient. This task focuses on evaluating whether the integrated basket offers all the necessary ingredients for a complete recipe. Unlike sequence-based approaches that use a `<stop>` token to signal the end of generation, ingredient data are unordered therefore there is no inherent position for a termination symbol to be correctly predicted.

For each recipe \(r_i \in R\), the whole ingredients set is denoted as a completed set with a positive label. At the same time, a subset of ingredients is randomly dropped to create an incomplete basket, which is labeled as negative class. In the S2SRec2, the representation of complete recipe - obtained after the set encoder and the subsequent prediction module - is passed through a fully connected layer that outputs a predicted probability \(p\) representing the likelihood that the augmented basket is complete. The training objective for this task is defined by the binary cross-entropy (BCE) loss:

\vspace{-1mm}
\begin{equation}
L_{BCE} = -\frac{1}{N} \sum_{i=1}^{N} \Big[ y_i \log(p_i) + (1-y_i) \log(1-p_i) \Big],
\end{equation}
\vspace{-2mm}

where \(y_i\) is the ground truth label with \(y_i=1\) indicating a complete basket and \(y_i=0\) an incomplete one and \(N\) is the number of training samples.

This completeness prediction task enables the model to learn nuanced representations of recipe completion, ensuring that the recommended ingredient complements the existing basket.
\vspace{-3mm}
\subsection{Joint Loss for Missing Ingredient Prediction and Set Completeness Prediction}

To integrate recipe knowledge into the completing task, we adopt a multi-task learning strategy, training both the missing ingredients prediction and set completeness prediction tasks in parallel. Each task is associated with its own specific output head, allowing the model to make complementary predictions for the main task while benefiting from shared latent representations. The overall objective is optimized using a combined loss function. 
\vspace{-2mm}
\begin{equation}
L_{joint}= \alpha \times L_{CE} + (1-\alpha) \times L_{BCE}.
\end{equation}
\vspace{-3mm}
A weighting factor $\alpha$ is introduced to balance these two objectives.

\subsection{Inference Process}
\vspace{-1mm}
\begin{figure*}[htbp]
\centering
\includegraphics[width=0.9\linewidth]{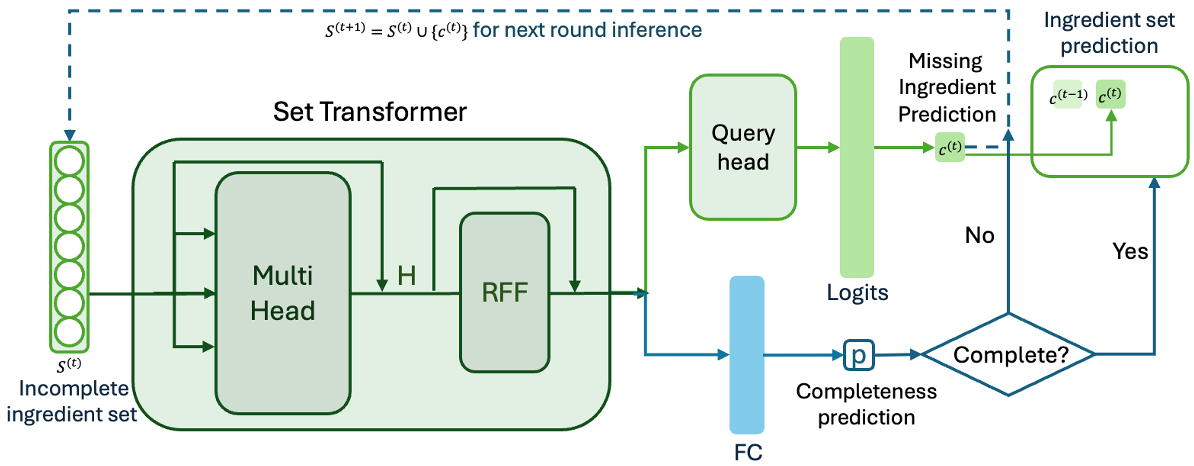}
\caption{
Sequential inference with S2SRec2. Check the completeness prediction, add the top predicted ingredient to input set; repeat until the completeness prediction is yes.
}
\label{fig:inference}
\end{figure*}
\vspace{-3mm}

During inference, the process begins by initializing an empty predicted ingredient set, \( c_p = \{\} \). For each inference round \( t \), the current basket of ingredients \( S^{(t)} \) is passed through S2SRec2, which outputs two predictions: a completeness probability \( p^{(t)} \) and a recommended ingredient \( c^{(t)} \). If \( p^{(t)} \leq 0.5 \), indicating that the basket is incomplete, the recommended ingredient \( c^{(t)} \) is added to the predicted set \( c_p \) and the basket is updated as

\vspace{-2mm}
\begin{equation}
S^{(t+1)} = S^{(t)} \cup \{ c^{(t)} \}.
\end{equation}
\vspace{-2mm}

This iterative process continues until the completeness probability exceeds 0.5 (i.e., \( p^{(t)} > 0.5 \)), at which point the basket is considered complete and the predicted ingredient set \( c_p \) is the final prediction. Figure \ref{fig:inference} demonstrates the inference procedure of S2SRec2.

\vspace{-3mm}
\section{Experiment and Results}


\vspace{-1mm}
\subsection{Data}
\vspace{-1mm}

Our recipes are sourced from an open-source dataset provided by food.com. To enhance data quality and reduce noise, we filter out recipes with fewer than 5 ingredients, as they often lack sufficient complexity for meaningful analysis. Additionally, we removed recipes with more than 15 ingredients to focus on the core structure of common recipes since 95\% of recipes have fewer than 15 ingredients. We also followed Kitchenette \cite{Park_2019} to remove ingredients whose occurrence count does not exceed 20. We end up with a final dataset of 141,782 recipes with 3,804 unique ingredients, allowing us to emphasize the most relevant and distinctive ingredients, while minimizing the influence of common, less informative ingredients. During training, each recipe is passed through randomly dropping up to three ingredients twice, which augments the data for learning. All ingredient embeddings are initialized using a pretrained BERT to project each ingredient into 768-dimensional vectors.
 Code is available at: \href{https://github.com/ycao21/S2Srec2}{https://github.com/ycao21/S2Srec2}.

\vspace{-1mm} 
\subsection{Evaluation}
\vspace{-1mm}

To evaluate set‐to‐set compatibility, we compare S2SRec2 against five baselines on the task of predicting missing ingredient(s) from a pool of 3,804 candidates given a partial basket. All methods share the same preprocessing and postprocessing pipeline to ensure fairness. The baselines are:
\vspace{-1mm}
\begin{itemize}
  \item \textbf{Logistic Regression (LR):} Bag-of-ingredients representation with pretrained embeddings, followed by a multi-label logistic regression classifier.
  \item \textbf{Vanilla Neural Network:} Pretrained Word2Vec ingredient embeddings are passed through two fully connected layers for multi-label ingredient prediction.
  \item \textbf{Bi-LSTM:} Arbitrary sequence structure on the unordered ingredient set to assess the impact of explicitly modeled order information
  \item \textbf{Kitchenette:} Adapted Siamese-network for ingredient pairing model, recontructed to accept multiple ingredients and output multi-label predictions.
  \item \textbf{Reciptor:} Set Transformer-based recipe representation learner with multi-task objectives, modified to produce multi-label ingredient outputs.
\end{itemize}
\vspace{-1mm}


We use Precision, Recall, and F1 score to evaluate the quality of the predicted ingredient set against the ground truth, and Mean Squared Error (MSE) to assess the stop‐head prediction. 

Let \(P\) be the predicted ingredient set and \(G\) the ground‑truth set. Then
\vspace{-2mm}
\[
\text{Precision}= \frac{|P \cap G|}{|P|},\quad
\text{Recall}= \frac{|P \cap G|}{|G|},\quad
\text{F1}= \frac{2\,\text{Precision}\times\text{Recall}}{\text{Precision}+\text{Recall}},\quad
\text{MSE}= \frac{1}{N}\sum_{i=1}^N \left(|P_i| - |G_i|\right)^2
\]
\vspace{-2mm}

These metrics are calculated using only the first $k$ predictions made by the model. If the model predicts more than $k$ ingredients, only the top $k$ are considered. This approach measures each model's ability to prioritize correct predictions early, focusing on the most relevant recommended ingredients.

The experiments are conducted uniformly applied to all models with Adam Optimizer with an initial learning rate of $10^{-4}$, 500 batch size and 30 training epochs. For multi-task variants, we tuned joint-loss weight $\alpha$ over \{0.2,0.4,0.6,0.8\} on the validation set, selecting $\alpha=0.6$ as the best trade-off between ingredient retrieval and stop prediction. All neural network implementations are developed using PyTorch 2.0. The experiments are carried out on a machine with two Nvidia T4 GPUs, providing a total of 60GB of memory.
\subsection{Results}

\begin{table*}[htbp]
\vspace{-3mm}
\caption{Performance comparison between S2SRec2 and baseline methods: set size prediction of baseline models are controlled by thresholds of multi-label task. The table demonstrates the performance of models with minimal size prediction error.}
\label{tab:performance}
\centering
\begin{tabular}{lccc ccc ccc}
\toprule
 & \multicolumn{2}{c}{Precision@k} & \multicolumn{2}{c}{Recall@k} & \multicolumn{2}{c}{F1@k} & \multicolumn{2}{c}{MSE@k} \\
\cmidrule(lr){2-3}\cmidrule(lr){4-5}\cmidrule(lr){6-7}\cmidrule(lr){8-9}
 & k=3 & k=5 & k=3 & k=5 & k=3 & k=5 & k=3 & k=5 \\
\midrule
LR & 0.0048 & 0.0194 & 0.0051 & 0.0284 & 0.0044 & 0.0197 & 1.7083 & 2.8845\\
Vanilla NN & 0.0022 & 0.0243 & 0.0019 & 0.0321 & 0.0019 & 0.0236 & 1.6894 & 2.9846  \\
Bi-LSTM & 0.0013 & 0.0102 & 0.0017 & 0.0176 & 0.0015 & 0.0120 & 1.6903 & 2.7203 \\
Kitchenette & 0.0043 & 0.0098 & 0.0052 & 0.0165 & 0.0043 & 0.011 & 1.6829 & 2.9829 \\
Reciptor & 0.0205 & 0.012 & 0.0271 & 0.018 & 0.0219 & 0.0126 & 1.5706 & 2.9578 \\
S2SRec2      & $\textbf{0.0331}$ & $\textbf{0.0307}$ & $\textbf{0.0377}$ & $\textbf{0.0399}$ & $\textbf{0.0334}$ & $\textbf{0.0323}$ & $\textbf{1.0331}$ & $\textbf{2.857}$ \\
\bottomrule
\end{tabular}
\end{table*}
\vspace{-1mm}

Table \ref{tab:performance} presents the performance of S2SRec2, compared with traditional and deep learning baselines on the set-to-set compatibility task. The baseline models are framed as a multi-label classification problem, where the number of predicted positive classes functions similarly to the ingredient completeness head and is controlled by thresholds. To ensure a fair comparison, baseline models are tuned to achieve the lowest set size prediction error, and the table demonstrates the performance of complementary ingredient prediction when all models reach their best set size prediction. Among the baselines, Logistic Regression and Vanilla Neural Network fall short in precision and recall due to their limited ability to capture complex ingredient interactions. The Bi-directional LSTM baseline, which imposes a sequential structure on unordered ingredient sets, performs worst overall, further supporting the importance of permutation-invariant modeling. Kitchenette and Reciptor show partial improvements but remain outperformed by S2SRec2 across all metrics. Notably, S2SRec2 also achieves the lowest MSE, indicating its superior ability to estimate set cardinality alongside accurate ingredient prediction.

Although precision and recall appear modest, this reflects the inherent difficulty of predicting exact ingredients from over 3,800 candidates—especially as multiple valid completions often exist (e.g., a sushi roll can validly include either rice vinegar or fresh vinegar, but only one appears as ground truth). Additionally, sequential inference in S2SRec2 compounds early prediction errors, making the task harder compared to multi-label methods predicting all ingredients simultaneously. Despite these challenges, S2SRec2 consistently outperforms baselines and introduces a practical stop mechanism. Unlike multi-label models that over-generate recommendations with a fix list of ranked candidates, S2SRec2 introduces an important practical capability: dynamically determines when to stop to avoid burdening users with additional filtering, providing coherent sets aligned with customer baskets. 

\begin{table*}[htbp]
\caption{Ablation study of S2SRec2}
\label{tab:ablation}
\centering
\begin{tabular}{lccc ccc ccc}
\toprule
 & \multicolumn{2}{c}{Precision@k} & \multicolumn{2}{c}{Recall@k} & \multicolumn{2}{c}{F1@k} & \multicolumn{2}{c}{MSE@k} \\
\cmidrule(lr){2-3}\cmidrule(lr){4-5}\cmidrule(lr){6-7}\cmidrule(lr){8-9}
 & k=3 & k=5 & k=3 & k=5 & k=3 & k=5 & k=3 & k=5 \\
\midrule
S2SRec2 w/o stop head & 0.0286 & 0.0202 & $\textbf{0.0436}$ & 0.0508 & 0.0331 & 0.0279 & 1.6729 & 9.6676\\
S2SRec2 multi-label & 0.0219 & 0.0213 & 0.0322 & $\textbf{0.0511}$ & 0.0248 & 0.0287 & 1.6764 & 9.2562  \\
Set encoder w/ stop head & 0.0304 & 0.0283 & 0.0348 & 0.0366 & 0.0307 & 0.0297 & 1.0407 & 2.8263 \\
Vanilla NN w/ stop head & 0.0182 & 0.0177 & 0.0134 & 0.0142 & 0.0141 & 0.0139 & 1.2528 & $\textbf{2.2459}$ \\
S2SRec2      & $\textbf{0.0331}$ & $\textbf{0.0307}$ & 0.0377 & 0.0399 & $\textbf{0.0334}$ & $\textbf{0.0323}$ & $\textbf{1.0331}$ & 2.757 \\
\bottomrule
\\
\end{tabular}
\end{table*}
\vspace{-5mm}

To better understand the contribution of each component in the proposed S2SRec2 model, we conduct an ablation study involving five variants, as shown in Table \ref{tab:ablation}. Removing the \textit{stop head} and instead truncating the output at a fixed length ($k=3$ or $k=5$) results in a drop of precision and F1. As the model is forced to output multiple ingredients regardless of confidence of completeness, it increases the likelihood of covering true positives, yielding the highest recall at first 3, but at the expense of precision and higher MSE due to over-generation. Replacing the stop head with a multi-label classification mechanism using a high threshold also degrades precision and F1, due to the same reason. When evaluating structural changes, we find that replacing the set-transformer with a mean-pooling-based set encoder while retaining the stop head leads to a significant drop in all missing ingredient prediction metrics. Replacing the set-transformer with a vanilla neural network shows the weakest performance across precision, recall, and F1, confirming the importance of permutation-invariant set modeling. Finally, the full S2SRec2 model achieves the best overall performance in precision, F1, and MSE, demonstrating that both the decoder-based set aggregation and the dedicated stop prediction mechanism are essential to accurate and controllable ingredient set completion. 
%


Table \ref{tab:qualitative_examples} presents qualitative examples comparing S2SRec2 with the best baseline model on the ingredient basket completion task based on the precision. S2SRec2 is able to predict key missing ingredients that align closely with the ground truth, while baseline models tend to over-generate unrelated ingredients.

\vspace{-3mm}
\begin{table*}[htbp]
\caption{Qualitative examples comparing S2SRec2 with the best baseline model on ingredient basket completion.}
\label{tab:qualitative_examples}
\centering
\resizebox{0.9\textwidth}{!}{ 
\begin{tabular}{p{1.5cm} p{5cm} p{2.25cm} p{2.25cm} p{2.5cm}}
\toprule
\textbf{Recipe Name} & \textbf{Input Ingredient Set} & \textbf{Ground Truth} & \textbf{S2SRec2 \quad Prediction} & \textbf{Baseline \quad \quad Prediction} \\
\midrule
Layered Taco Dip & Lettuce, taco seasoning, cheddar cheese, cream cheese, rom tomato, guacamole, black olive & Refried bean & Refried bean & Mayonnaise, red onion, avocado, celery, ... \\
\midrule
Texas Armadillo Balls & Black pepper, red pepper flake, chicken broth, jalapeno pepper, celery, cayenne pepper, habanero pepper, breadcrumb, chicken soup & Cornbread, green pepper & Cornbread, pepper cheese & Chicken breast, Worcestershire sauce, bay leaf, ... \\
\midrule
Thai Tenderloins & Fresh cilantro, peanut butter, garlic, tabasco sauce, soy sauce, fresh ginger & Pork tenderloin, chicken stock, orange juice & Pork tenderloin, pineapple juice & Honey, cornstarch, cayenne pepper, celery, ... \\
\bottomrule
\end{tabular}
}
\end{table*}
\vspace{-3mm}

\vspace{-2mm}
\section{Conclusion}
\vspace{-1mm}
In this paper, we introduced S2SRec2, a set-based recommendation framework for ingredient basket completion that bridges the gap between grocery baskets and culinary experiences. Our approach leverages a Set Transformer to capture inter-dependencies among ingredients within the basket, followed by two complementary tasks. The first task employs a learnable query combined with cross-entropy loss to predict missing ingredients, ensuring precise retrieval of complementary items. The second task uses binary cross-entropy loss to assess basket completeness, determining whether the union of existing and predicted ingredients forms a complete recipe. By jointly optimizing both tasks, S2SRec2 not only enhances ingredient compatibility but also ensures coherent ingredient recommendations.

While S2SRec2 focuses on recommending complementary ingredients to complete a basket, it is promising to be extended to multiple real-world e-commerce settings. For example, the representation of a completed basket can be obtained from S2SRec2 and used to identify and surface full recipes that match the completed basket. Furthermore, integrating user-specific signals such as past purchase history and dietary restrictions into the set representation could personalize both ingredient and recipe recommendations. In future work, we will deploy S2SRec2 in production environment at scale and conduct online evaluations under live user traffic. Collectively, S2SRec2 can evolve from a standalone complement recommender into a fully personalized, interactive recipe discovery and completion engine for grocery e-commerce.



\vspace{-2mm}
\section*{Declaration on Generative AI}
\vspace{-2mm}
During the preparation of this paper, the author(s) used ChatGPT (GPT‑4) for language-quality assistance (grammar checking, rephrasing, and minor stylistic edits). The author(s) reviewed, edited, and approved all content and take full responsibility for the final publication.
\bibliography{bibfile}


\appendix
\section{Set Transformer}
\label{appendix}

The Set Transformer consists of an encoder and a decoder, each serving a distinct purpose. 
The encoder processes each element independently by performing self-attention among the elements in the set. This results in an output set of equal size, where each element's representation is enriched by its relationship with the other elements. Set Encoder consists of two Induced Set Attention Blocks, which are a variation of Set Attention Block that reduces the computing complexity of large sets. Both blocks are based on Multi-head Attention, which computes attention scores of different projections of the input vector.

Specifically, given a set of $n \times d_q$ dimensioned query vectors $Q \in R^{n \times d_{q}}$ and their corresponding key-value pairs $K \in R^{n \times d_{k}}$ and $V \in R^{n \times d_{v}}$  as input, the attention function is defined as 
\begin{equation}
Attention(Q,K,V)=softmax(\phi(QK^{T})V),
\end{equation}
where $\phi=\frac{1}{\sqrt{d_k}}$ is the scale function. The attention function computes the dot product of $Q$ and $K$ and applies the scaling factor to output a weighted sum of $V$ values, assigning a higher weight when the dot product of the query and key is large. 

Instead of single-head attention, which only captures inter-element relationships in a single space, multi-head Attention \cite{vaswani2017attention} is introduced to generate different outputs in multiple subspaces:
\begin{equation}
MultiHead(Q,K,V)= (A_1 \oplus ... \oplus A_k ),
\end{equation}
\begin{equation}
A_i= Attention(QW_{i}^{Q},KW_{i}^{K},VW_{i}^{V} ).
\end{equation}

The Multi-head Attention Blocks (MAB) in Set Transformer are built using the multi-head Attention in combination with layer normalization.
\begin{equation}
MAB(X,Y)  = LayerNorm(H + RFF(H)),
\end{equation}
where $H = LayerNorm(X + Multihead(X,X,X))$.

To improve the training efficiency, the Set Transformer proposed the Induced Set Attention Blocks (ISAB) \cite{lee2019set}, which employ a trainable inducing vector $K$ fed twice in a MAB, significantly reducing the computing complexity especially for large sets.
\begin{equation}
ISAB = MAB(X,MAB(K,X)),
\end{equation}

The decoder summarizes the set via pooling for downstream tasks. Rather than using dimension-wise mean pooling, the Set transformer applies a row-wise feedforward layer (RFF) in conjunction a Multi-head Attention-based Pooling layer (PMA) on a set of $m$ trainable vectors $S$. Given a set of vectors $X \in R^{n \times d}$, the entire process can be expressed as
\begin{equation}
PMA = MAB(ISAB(SAB(X)),RFF(S)),
\end{equation}
\begin{equation}
SAB = MAB(X,X).
\end{equation}

\end{document}